# An Efficient Watermarking Algorithm to Improve Payload and Robustness without Affecting Image Perceptual Quality

Er.Deepak Aggarwal, Er.Sandeep Kaur and Er.Anantdeep

**Abstract**— Capacity, Robustness, & Perceptual quality of watermark data are very important issues to be considered. A lot of research is going on to increase these parameters for watermarking of the digital images, as there is always a tradeoff among them. . In this paper an efficient watermarking algorithm to improve payload and robustness without affecting perceptual quality of image data based on DWT is discussed. The aim of the paper is to employ the nested watermarks in wavelet domain which increases the capacity and ultimately the robustness against attacks and selection of different scaling factor values for LL & HH bands and during embedding not to create the visible artifacts in the original image and therefore the original and watermarked image is similar.

**Index Terms**—Watermarked, Perceptual, Matlab,

—————————— ◆ ——————————

## 1 INTRODUCTION

To increase payload the one watermark is embedded in the other i.e nested watermark is used. Two different visual watermarks are used, nested and thus the resulting watermark is embedded in lower (LL) & high frequency (HH) bands based on the optimal selection of different scaling factors for each band. The DWT coefficients in lower frequency band are larger as compared to the higher frequency bands. Therefore the value of scaling factor is kept larger in lower frequency band and lower in high frequency band. The scaling factors are chosen such invisibility and quality of extracted watermark is maintained. Visual quality of extracted watermarks is measured by the Similarity Ratio (SR) between compared images.In this paper we are giving a new image watermarking method. The embedding and extraction of watermark is based on discrete wavelet transform. Matlab 6 [28] is used to implement all the coding related to digital image processing .It is a non-blind watermarking method.

## 2. ALGORITHMS

**2 .1 Algorithm to embed one watermark into other watermark:**

Inputs: Primary watermark, secondary watermark image.
 Steps:
 1) Read the primary visual watermark image.
 2) Decompose the primary watermark image into cap1, chp1, cvp1, cdp1 bands using daubachesis filter.
 3) Read the secondary visual watermark image.
 4) Add the secondary watermark into the horizontal DWT coefficient (chp1) of Primary watermark.
 5) Apply the IDWT to get the nested watermark image.
 6) Calculate the PSNR & MSE of nested watermark with original primary watermark image.
Output: Nested watermark image

**2.2 Algorithm to embed nested watermark into cover image:**

Inputs: Cover image, nested watermark image
Steps:
 1) Read the cover image & nested watermark image.
 2) Apply DWT to cover image to obtain approximation, horizontal, vertical, diagonal DWT coefficients i.e. ca1, ch1, cv1, cd1.
 3) Modify the approximztion DWT coefficient by adding the nested watermark image as in equation
    a. Ca1(i,j)=ca1(i,j)+($\alpha$*nested watermark)
    b. Where Ca1 & ca1 are the modified & original approximation coefficients and $\alpha$ is a scaling value as set to 0.04.
 4) Modify the diagonal DWT coefficient by adding the nested watermark image as in the equation
    a. Cd1(i,j)=cd1(i,j)+( $\alpha$*nested watermark)
 5) Where Cd1 & cd1 are the modified & original diagonal coefficients of cover image and $\alpha$ is set to value 0.01.

6) Apply the inverse DWT to obtain the watermarked cover image.
7) Calculate the PSNR & MSE of watermarked cover image with original cover image.

Output: Watermarked cover image.

**2. 3 Algorithm for Watermark Extraction:**

Inputs: Original cover image, Watermarked cover image or Attacked Watermarked cover image.

1) Apply two-dimensional DWT, to obtain the first level decomposition of the watermarked cover image i.e. c1a1, c1h1, c1v1, c1d1
2) Extract the watermark from approximation & diagonal DWT coefficient (c1a1 & c1d1) as per the equation
   $LL_w = (c1a1-ca1)/\alpha$; where $\alpha=0.04$
   $HH_w = (c1d1-cd1)/\alpha$; where $\alpha=0.01$
3) Calculate the visual quality of extracted watermark by the Similarity Ratio (SR) between compared images. SR= S/(S+D), where S denotes the number of matching pixel values in compared images, and D denotes the number of different pixel values in compared images.
4) Apply the set of possible attacks to watermarked image to get the attacked image. Calculate for each attack the PSNR of Original and attacked image.
5) Apply two-dimensional DWT, to obtain the first level decomposition of the Attacked image i.e. c2a2, c2h2, c2v2, c2d2.
6) Extract the watermark from approximation & diagonal DWT (i.e. c2a2 & c2d2) coefficient of attacked image as per the equation
   $LA_w = (c2a2-ca1)/\alpha$; where $\alpha=3$
   $HA_w = (c2d2-cd1)/\alpha$; where $\alpha=1$
7) Calculate the visual quality of extracted watermark from attacked image by the Similarity Ratio (SR) between compared images. SR= S/(S+D).

Outputs: Extracted Watermarks from approximation & diagonal coefficients of watermarked cover image & Attacked cover image

## 3 EXPERIMENTAL RESULTS

In our experimental results Lena cover image of 512*512 size, Primary and Secondary watermark of 64*64 size is used. Cover image is subjected to the watermark embedding and extraction. We measure the quality of watermarked images in terms of PSNR (Peak Signal to Noise Ratio) and MSE (Mean Square Error). In ideal case PSNR should be infinite and MSE should be zero. But it is not possible for watermarked image. So, large PSNR and small MSE is desirable.

$$PSNR_{dB} = 10 \log_{10} \left( \frac{MAX^2}{MSE} \right)$$

The subjective evaluation of extracted watermark is based on Similarity Ratio SR.

$$SR = \frac{S}{S+D}$$

Where S denotes the number of matching pixel values in compared images, and D denotes the number of different pixel values in compared images.

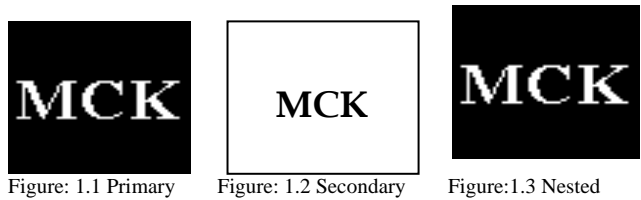

Figure: 1.1 Primary    Figure: 1.2 Secondary    Figure:1.3 Nested

Figure 1: Watermark Images

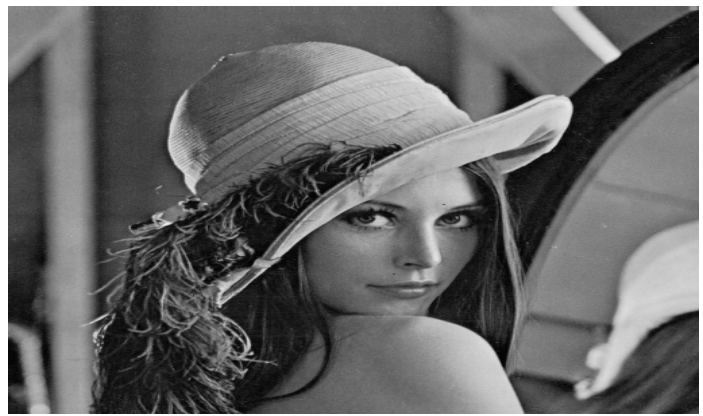

Figure 2: Original Image

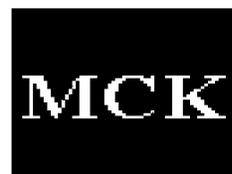 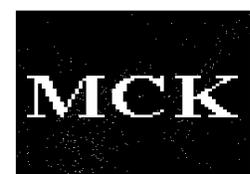

**Figure 3.1: LL (SR 9226)**        **Figure3.2: HH (SR 0.9181)**

Figure 3: Extracted watermarks

Table 1: PSNR1 & PSNR2

| Cover Image | Primary Watermark Image | Secondary Watermark Image | PSNR1 | MSE1 | PSNR2 | MSE2 |
|---|---|---|---|---|---|---|
| Lena 512*512 | dmg2.tif (64 * 64) | dmg1.tif (64*64) | 42.14 | 6.10e-005 | 54.18 | 3.81e-006 |

PSNR1 – PSNR of Nested watermark with original Primary Watermark.

MSE1 – MSE of Nested watermark with original Primary Watermark.

PSNR2 – PSNR of gray scale cover image after embedding Nested watermark.

MSE2 – MSE of gray scale cover image after embedding Nested watermark.

**3.1 Capacity Increase Results**

In our Watermarking technique the embedding capacity is more than normal Watermarking because here watermark nesting is used.

Table 2: Capacity Increase Results

| Cover Image | Capacity of bits without watermarking | Capacity of bits with watermark |
|---|---|---|
| Lena | 4096 | 8192 |

Table 3: Robustness Improve Results

| Cover Image | Nested Watermark | Type of attack | PSNR | SR | Extracted band |
|---|---|---|---|---|---|
| Lena Watermark.tif (512*512) (64*64) | | Intensity Adjustment | 29.57 | 0.8749 | HH |
| | | Gamma Correction | 23.30 | 0.8680 | HH |
| | | Histogram Equalization | 30.66 | 0.7269 | HH |
| | | Low Pass Filter/Blur | 21.31 | 0.8803 | LL |
| | | Resizing | 54.18 | 0.7440 | LL |
| | | Gaussian Noise | 54.18 | 0.7707 | LL |
| | | High Pass Filter/ Sharp | 19.22 | 0.7582 | LL |
| | | JPEG Compression | 42.14 | 0.9226 | LL |

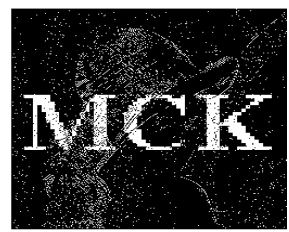 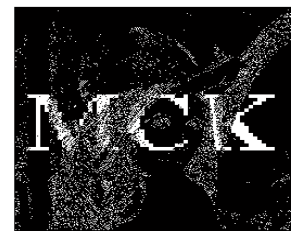

Figure 4.1: Intensity Adjustment ( [0;0.8],[0;1] )      Figure 4.2: Gamma Correction (1.5)

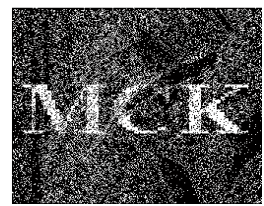 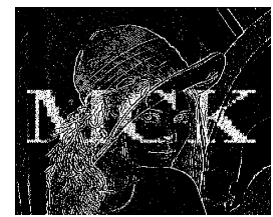

Figure 4.3: Histogram Equalization      Figure 4.4: Low Pass Filter (3, 3)

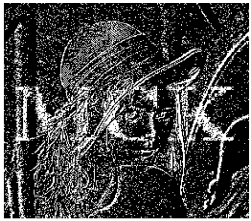

Figure 4.5: Resizing
(512,256,512)

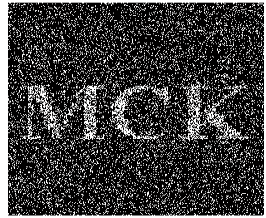

Figure 4.6: Gaussian Noise
(mean=0, variance 0.001)

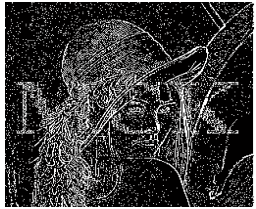

Figure 4.7: High Pass Filter [0.6]

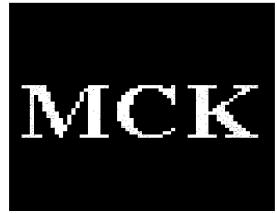

Figure 4.8: JPEG Compression

Figure 4: Attacks

## 4. CONCLUSION

This paper presents a non-blind watermarking technique that uses watermark nesting at level 1 DWT decomposition. Nesting means it embeds an extra watermark into the main watermark and then embeds the main watermark into the cover image. Proposed watermarking technique has following advantages:

(1) Proposed watermarking embeds more no of bits into cover image and thereby improves the payload as compared to single watermark embedding.
(2) Watermark embedding in the LL band is most resistant to JPEG compression, blurring, Sharpening, adding Gaussian noise and Resizing.
(3) Watermark embedding in the HH band is most resistant to histogram equalization, intensity adjustment, Gamma Corrrection.
(4) Recovered watermarks quality under the above attacks is assessed based on subjective evaluation of similarity Ratio(SR). In all attacked cases the SR value is more than 0.7.
(5) Robustness is improved in all the attacked watermarked images as PSNR shows a good value against the original watermarked image.
(6) The scaling factors are chosen such as invisibility and quality of extracted watermark is maintained.

## 5. FUTURE WORK

Most of the research is going in watermarking the text, audio, video data for copyright protection and authentication of electronic documents and media. Watermarking is the necessity for digital images as these are available at Internet without any cost, which needs to be protected. The watermarking technique that is given in this paper can be further improved to increase the hiding capacity and Robustness for RGB and the Indexed images.

## REFERENCES

[1] W. Stallings, Cryptography and Network Security, second edition. Prentice Hall, 1998.
[2] Dr. Martin Kutter and Dr. Frederic Jordan, "Digital Watermarking Technology", AlpVision, Switzerland, pp 1 – 4.
[3] Chun-Shien Lu, Multimedia Security: "Stegranograhy and Digital WatermarkingTechniques for Protection of Intellectual Property", Idea Group Inc, 2005, USA.
[4]. "Digital Watermarking" available at
 http://en.wikipedia.org/wiki/Digital_watermarking
[5] Saraju Prasad Mohanty, "Watermarking of Digital Images", Submitted at Indian Institute of Science Bangalore, pp. 1.3 – 1.6, January 1999.
[6] A. Nikolaidis, S. Tsekeridou, A. Tefas, V Solachidis, "A Survey On Watermarking Application Scenarios And Related Attacks", IEEE international Conference on Image Processing, Vol. 3, pp. 991-993, Oct. 2001.(applications)
[7] Alper Koz, "Digital Watermarking Based on Human Visual System", TheGraduate School of Natural and Applied Sciences, The Middle East TechnicalUniversity, pp 2 – 8, Sep 2002.
[8] J.J.K.O. Ruanaidh, W.J.Dowling, F.M. Boland, "Watermarking Digital Images for Copyright Protection", in IEE ProcVis. Image Signal Process. Vol. 143, No. 4, pp 250 - 254. August 1996.
[9] D. H. D. Kundur, Digital Watermarking for Telltale Tamperproofing and Au-thentication, vol. 87(7), pp. 1167{1180. Proceedings of the IEEE Special Issueon Identi¯cation and Protection of Multimedia Information, Jul. 1999.
[10] Franco A. Del Colle_ Juan Carlos G´omez "DWT based Digital Watermarking Fidelity and Robustness Evaluation" JCS&T Vol. 8 No. 1,April 2008.
[11] Ali Al-Haj "Combined DWT-DCT Digital Image Watermarking" Journal of Computer Science 3 (9): 740-746, 2007.
[12] A White paper on "Digital Watermarking: A Technology Overview", Wipro Technologies, pp. 2 – 8. Aug. 2003.
[13] Peining Tao and Ahmet M. Eskicioglu "Robust multiple watermarking scheme in the Discrete Wavelet Transform domain" Optics East 2004, Philadelphia.
[14] I. J. Cox, M. L. Miller, and J. A. Bloom, Digital Watermarking, Morgan Kaufmann Publishers, 2002.
[15] S. P. Mohanty, K. R. Ramakrishnan, and M. S. Kanakanhalli, "A Dual Watermarking Technique for Images", in Proceedings of the 7th ACM International Multimedia Conference (ACMMM) (Vol. 2), pp.49-51, 1999.
[16].. Saraju P. Mohanty, Bharat K. Bhargava "Invisible Watermarking Based on Creation and Robust Insertion-Extraction of Image Adaptive Watermarks" ACM Journal , Vol. V, No. N, Pages 1–24, February 2008.

[17] Rakesh Dugad, Krishna Ratakonda, and Narendra Ahuja, "A new waveletbased scheme for watermarking images," in Proceedings of the IEEE International Conference on Image Processing, ICIP '98, Chicago, IL, USA, October 1998.
[18] T. Palfner, M. Schlauweg and E. Müller, "A Secure Semi-fragile Watermarking Algorithm for Image Authentication in the Wavelet Domain of JPEG2000" The Second International Conference on Innovations in Information Technology (IIT'05). Sviatoslav Voloshynovskiy, Shelby Pereira, and Thierry Pun, "Attacks on digital watermarks: Classification, Estimation-based attacks, and benchmarks," IEEE Communcations Magazine, August 2001.
[19] "MATLAB - The Language of Technical Computing" available at http://www.mathworks.com/access/helpdesk/matlab/getstart.pdf
[20] Joseph J. K. Ruanaidh and T. Pun, "Rotation, scale and translation invariantdigital image watermarking," Proc. ICIP'97, IEEE Int. Conf. Image Processing, Santa Barbara, CA, Oct. 1997, pp. 536-539.
[21] Frank Hartung, Martin Kutter, "Multimedia Watermarking Techniques", Proceedings of The IEEE, Vol. 87, No. 7, pp. 1085 – 1103, July 1999.
[22] Marco Corvi and Gianluca Nicchiotti, "Wavelet-based image watermarking for copyright protection," in Scandinavian Conference on Image Analysis SCI '97, Lappeenranta, Finland, June 1997.
[23] Jong Ryul Kim and Young Shik Moon, "A robust wavelet-based digital watermark using level-adaptive thresholding," in Proceedings of the 6th IEEE International Conference on Image Processing ICIP '99, page 202, Kobe, Japan, October 1999.
[24] Xiang-Gen Xia, Charles G. Boncelet, and Gonzalo R. Arce, "Wavelet transform based watermark for digital images," Optics Express, 3 pp. 497, December 1998.
[25] Liehua Xie and Gonzalo R. Arce, "Joint wavelet compression and authentication watermarking," in Proceedings of the IEEE International Conference on Image Processing, ICIP '98, Chicago, IL, USA, 1998.
[26] "Fundamentals of Wavelets" available at
http://documents.wolfram.com/applications/wavelet/index2.html
[27] Amara Graps, "An Introduction to Wavelets", in IEEE Computer Science and Engineering, vol. 2, num. 2, pp. 50-59, 1995.

**Er. Deepak Aggarwal**, presently working as Lecturer in the Department CSE/IT of BBSB Engineering College, Fatehgarh Sahib, Punjab (INDIA). He is having a total teaching experience of about 7 years & he has done Master of Technology from Punjab Technical University. His major research interests include DIP and performance evaluation of networks. Also Deepak Aggarwal is having to his credit about 8 publications in various National and International Conferences and Journals.

**Er. Sandeep Kaur**, presently working as Lecturer in the Department CSE/IT of BBSB Engineering College, Fatehgarh Sahib, Punjab (INDIA). She is master of engineering in computer science & engineering from Thapar University Patiala in 2009. Her major research interests include parallel computing and performance of optical multistage interconnection networks. Also Sandeep kaur is having 10 publications in various National and International Conferences.

**Er. Anantdeep**, presently working as Lecturer in the Department of CSE/IT, BBSB Engineering College, Fatehgarh Sahib. She is M-Tech in ICT from Punjabi university Patiala. Her major research interests include mobile zigbee networks . Also Anantdeep is having 4 publications in various National and International Conferences.